\pdfoutput=1

\documentclass[11pt]{article}

\usepackage{EMNLP2022}

\usepackage{times}
\usepackage{latexsym}

\usepackage[T1]{fontenc}

\usepackage[utf8]{inputenc}

\usepackage{microtype}

\usepackage{ascii}
\usepackage{hyperref}
\usepackage{amsmath}
\usepackage{amsfonts}
\usepackage{algorithm}
\usepackage[noend]{algpseudocode}
\usepackage{bm}
\usepackage{bbm}
\usepackage{booktabs}
\usepackage{balance}
\usepackage{color,colortbl}
\usepackage{CJKutf8}
\usepackage{graphicx} 
\usepackage{graphics}
\usepackage{multirow}
\usepackage{makecell}
\usepackage{paralist}
\usepackage{subfig}
\usepackage{textcomp}
\usepackage{threeparttable}
\usepackage[normalem]{ulem}
\usepackage{arydshln}
\usepackage{enumitem}

\usepackage{cleveref}
\crefformat{section}{\S#2#1#3} 

\definecolor{jred}{RGB}{196, 38, 11}
\definecolor{jblue}{RGB}{41, 52, 190}
\definecolor{jgreen}{RGB}{18, 141, 21}

\definecolor{wxjiao}{RGB}{18, 21, 141}

\title{Tencent's Multilingual Machine Translation System for WMT22 Large-Scale African Languages}

\author{
Wenxiang Jiao  \enspace Zhaopeng Tu \enspace Jiarui Li \enspace Wenxuan Wang \enspace Jen-tse Huang \enspace Shuming Shi \\
Tencent AI Lab \\  {\asciifamily \normalsize \tt \{joelwxjiao,zptu,ultrali,jwxwang,jentsehuang,shumingshi\}@tencent.com} \\
}

\begin{document}
\maketitle
\begin{abstract}
This paper describes Tencent's multilingual machine translation systems for the WMT22 shared task on Large-Scale Machine Translation Evaluation for African Languages. We participated in the \textbf{constrained} translation track in which only the data and pretrained models provided by the organizer are allowed.
The task is challenging due to three problems, including the absence of training data for some to-be-evaluated language pairs, the uneven optimization of language pairs caused by data imbalance, and the curse of multilinguality. 
To address these problems, we adopt data augmentation, distributionally robust optimization, and language family grouping, respectively, to develop our multilingual neural machine translation~(MNMT) models.
Our submissions won the \textbf{1st} place on the blind test sets in terms of the automatic evaluation metrics.\footnote{Codes, models, and detailed competition results are available at \url{https://github.com/wxjiao/WMT2022-Large-Scale-African}.}
\end{abstract}

\section{Introduction}
\label{sec:introduction}

Multilingual neural machine translation (MNMT) aims to translate between multiple language pairs with a unified model~\cite{Johnson2017GooglesMN}. It is appealing due to the model efficiency, easy deployment, and knowledge transfer between high resource languages and low resource languages. Hence, MNMT has attracted more and more attention from both academia and industry. To improve the performance of MNMT models, previous researchers have proposed various approaches on advanced model architectures~\cite{Sen2019MultilingualUN, Zhang2021ShareON}, training strategies~\cite{Wang2020BalancingTF,Wang2020MultitaskLF}, and data utilization~\cite{Siddhant2020LeveragingMD, wang2022UncertZS}. 
In addition, industrial companies have released massive multilingual pretrained models~\cite{Tang2021MultilingualTF} and large-scale multilingual translation models~\cite{Fan2021BeyondEM,team2022NoLL} to facilitate translation among hundreds of languages.  
However, existing efforts on MNMT for African languages are not sufficient due to the lack of high quality and standardized evaluation benchmarks.

In this paper, we build a system integrating several advanced approaches for WMT22 Large-Scale Machine Translation Evaluation Task, which involves a set of 24 African languages.
We participated in the Constrained Translation track, where only the data provided by the organizer are allowed.
This task is challenging due to three potential problems:
\begin{itemize}[leftmargin=10pt]
    \item The absence of training data for some to-be-evaluated language pairs;
    \item The uneven optimization of language pairs due to data imbalance;
    \item The curse of multilinguality in MNMT models caused by the hundreds of language pairs.
\end{itemize}

For the first problem, we adopt data augmentation techniques to construct synthetic data for the language pairs without parallel training data~(\cref{sec:data-augmentation}). Specifically, we use back-translation~\cite{sennrich2016BT} and self-training~\cite{jiao2021self}, and attach a special tag to the synthetic side of the data.
For the second issue, we utilize distributionally robust optimization~(DRO) method~\cite{oren2019DROforLM,zhou2021DROforMNMT} to balance the optimization process for different translation directions~(\cref{sec:distributionally-robust-optimization}).
For the third issue, we isolate the potential conflicts between language pairs by language family grouping and finetune a model for each language group~(\cref{sec:language-family-grouping}).

\begin{table*}[t!]
\small
\setlength{\tabcolsep}{4pt}
\centering
\caption{Information of language groups and the corresponding language pairs. We include additional 36 language pairs (\textbf{bolded}) to help the long-tail languages.
}
\vspace{-5pt}
\begin{tabular}{l p{14cm}}
\toprule
\bf Group & \multicolumn{1}{c}{\bf Language Pairs} \texttt{(73){\bf\texttt{(117)}}} \\
\midrule
\textsc{EngC} & afr-eng,amh-eng,eng-fra,eng-fuv,eng-hau,eng-ibo,eng-kam,eng-kin,eng-lug,eng-luo,eng-nso,eng-nya,eng-orm,eng-sna,eng-som,eng-ssw,eng-swh,eng-tsn,eng-tso,eng-umb,eng-xho,eng-yor,eng-zul,\texttt{(23)},{\bf eng-lin, eng-wol,\texttt{(25)}} \\[1ex]
\textsc{FraC} &  fra-kin,fra-lin,fra-swh,fra-wol,\texttt{(4)},{\bf amh-fra,fra-kam,fra-lug,fra-luo,fra-orm,fra-umb,\texttt{(10)}} \\[1ex]
\textsc{SSEA} & {\bf afr-nso},{\bf afr-sna},afr-ssw,afr-tsn,afr-xho,afr-tso,afr-zul,nso-sna,nso-ssw,nso-tsn,nso-xho,nso-tso,nso-zul,sna-ssw,sna-tsn,sna-xho,sna-tso,sna-zul,ssw-tsn,ssw-xho,ssw-tso,ssw-zul,tsn-xho,tsn-tso,tsn-zul,tso-xho,tso-zul,xho-zul, \texttt{(28)} \\[1ex]
\textsc{HCEA} & {\bf amh-luo},amh-orm,amh-som,amh-swh,luo-orm,luo-som,luo-swh,orm-som,orm-swh,som-swh, \texttt{(10)} \\[1ex]
\textsc{NGG} & fuv-hau,fuv-ibo,fuv-yor,hau-ibo,hau-yor,ibo-yor, \texttt{(6)} \\[1ex]
\textsc{CA} & {\bf kin-lin},kin-lug,kin-nya,kin-swh, {\bf lin-lug},{\bf lin-nya},{\bf lin-swh},lug-nya,lug-swh,nya-swh, \texttt{(10)} \\[1ex]
\textsc{Other} &\bf fuv-kin,fuv-nya,fuv-som,fuv-zul,kam-nya,kam-sna,kam-som,kam-swh,kam-tso,kam-zul,kin-yor,lug-sna,lug-zul,luo-nya,luo-sna,luo-zul,nya-umb,nya-yor,sna-umb,sna-yor,som-wol,som-yor,swh-umb,swh-yor,tso-yor,umb-zul,xho-yor,yor-zul,\texttt{(28)}
\\[1ex]
\bottomrule
\end{tabular}
\label{tab:group-langpairs-1}
\vspace{-5pt}
\end{table*}

Experimental results show that our system can significantly improve the performance of vanilla MNMT models, from 15.50 to 17.95 BLEU points~(\cref{sec:results}).
Extensive analysis suggests that data augmentation could be harmful to the translation performance if used for training the final models directly, due to the error-prone synthetic sentence pairs.
Instead, we utilize the resulting MNMT models as pretrained models to further finetune on clean datasets for the final models.
The DRO technique is very effective in improving the translation quality across all language pairs, particularly on the dominant languages (e.g., eng and fra), which also calls for an improved DRO to benefit more on other languages.
As for language family grouping, it especially improves the translation quality on one-to-many translations, which demonstrates its effectiveness in alleviating the curse of multilinguality issue.
Finally, our submission won the \textbf{1st} place in the official evaluation in terms of the automatic evaluation metrics.

\section{Data}
\label{sec:data}

In this section, we present the details of our data preparation.

\subsection{Language Pairs}
\label{sec:data-language-pairs}

We utilize all available datasets from the official website (including those from the Data Track participants)\footnote{\url{https://www.statmt.org/wmt22/large-scale-multilingual-translation-task.html}}, which provide either monolingual or parallel sentences.
According to the evaluation instruction, we group the language pairs into 7 groups, namely, English-Centric~(\textsc{EngC}), French-Centric~(\textsc{FraC}), South/South East Africa (\textsc{SSEA}), Horn of Africa and Central/East Africa (\textsc{HCEA}), Nigeria and Gulf of Guinea (\textsc{NGG}), Central Africa (\textsc{CA}), and Other related pairs (\textsc{Other}), to train the MNMT models.
Details are listed in Table~\ref{tab:group-langpairs-1}.

We consider three subsets of language pairs for training different models:
\begin{itemize}[leftmargin=10pt]
    \item \textbf{Base-146}: We train the \textsc{Transf-Deep}~(\cref{sec:settings}) models on the to-be-evaluated language pairs in the first 6 groups, as well as the English-French~(i.e., eng-fra) pair. In total, there are 81 language pairs but only 73 of them are provided with bitext data, which cover 146 translation directions (i.e., including both forward and backward).
    \item \textbf{Large-234}: The main issues of \textbf{Base-146} are that, some to-be-evaluated language pairs~(e.g., afr-nso) are missing in the training data and some languages are heavily long-tailed due to the imbalanced choice of language pairs. To alleviate these issues, we extend another 36 language pairs for the long-tail languages and construct synthetic data for all the language pairs in \textsc{EngC}, \textsc{SSEA}, \textsc{HCEA}, \textsc{NGG} and \textsc{CA}, which enables the training on 234 translation directions. We use these language pairs to train the \textsc{Transf-DWide}~(\cref{sec:settings}) models.
    \item \textbf{Eval-106}: The official evaluation includes 100 translation directions\footnote{\url{https://docs.google.com/document/d/11NYyJpJ4nhNIllwmF5kjkqfkaaEzXNU-CCO5E64MRDU/edit}}, which were notified at the later stage of the competition.
    We focus on these directions by finetuning the \textsc{Transf-DWide} models on these directions. To ensure the data amount of each language, we include all \textsc{EngC} directions, making the final 106 directions.
\end{itemize}

\begin{figure*}[t!]
    \centering
    \includegraphics[height=0.31\textwidth]{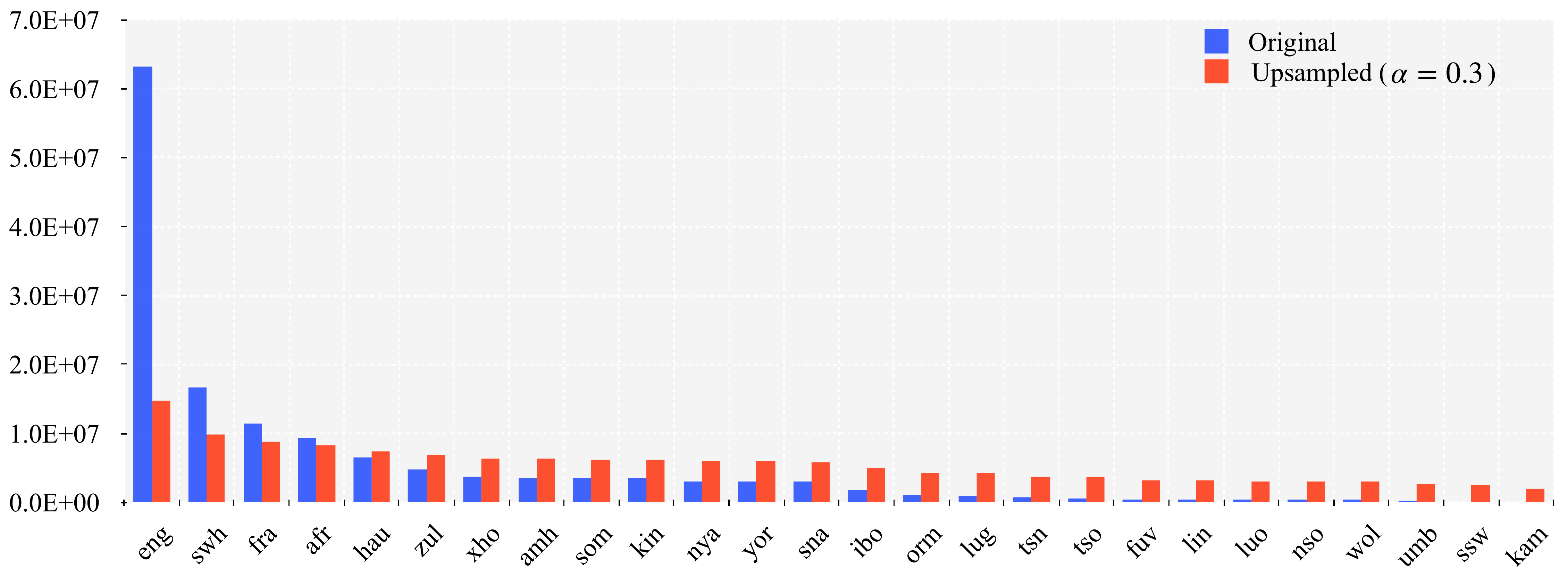}
    \caption{Number of sentences in each language and the upsampled distribution with the smoothing rate of $\alpha=0.3$.}
    \label{fig:langs_distribution}
    \vspace{-5pt}
\end{figure*}

\subsection{Data Preprocessing}
\label{sec:data-preprocessing}

We preprocess the raw and potentially noisy data by four steps, namely, reformatting, deduplication, language detection, and length limitation. Details are elaborated as below.

\paragraph{Reformatting.}
The raw data is stored in various alignment structures, including HTML, JSON, and special spacing. To reduce data noise, we reformat all data into a line-by-line tight structure and realign those missing paired ones.

\paragraph{Deduplication.}
We remove the duplicated sentences (pairs) in each monolingual and parallel dataset.
This aims to reduce information redundancy so that the MNMT models can be trained more efficiently.

\paragraph{Language Detection.}
Previous studies suggest that incorrect languages in training data induce translation uncertainty for both bilingual~\cite{ott2018analyzing} and multilingual~\cite{wang2022UncertZS} NMT models.
Therefore, we conduct language detection for all the datasets using \texttt{langid}\footnote{\url{https://github.com/saffsd/langid.py}}. 
Since \texttt{langid} neither supports all African languages nor performs well when distinguishing two African languages, we adopt a simplified strategy: for the African datasets, we remove sentences (pairs) that are identified as languages other than the 24 designated African languages. In other words, sentences (pairs) in one African language identified as another African language are also considered valid. For English and French datasets, we strictly restrict the correct languages as themselves, i.e., English and French, respectively.

\paragraph{Length Limitation.}
After multilingual tokenization, we conduct further filtering and retain sentence pairs with tokens between  4~\cite{Wu2019ExploitMonoScale} and 512~\cite{Yang2021MicrosoftWMT}, as well as the length ratio below 3.

\subsection{Multilingual Tokenization}
\label{sec:data-tokenize}

To tokenize the multilingual sentences, we follow~\citep{conneau2020xlmr} to train a Sentence Piece Model (SPM) and apply it directly on the preprocessed text data for all languages.
However, the distribution of data across languages is heavily long-tailed, as shown in Figure~\ref{fig:langs_distribution}.
To balance the vocabulary bandwidth between high-resource and low-resource languages, we follow~\newcite{conneau2020xlmr} to upsample the low-resource languages with a smoothing rate of $\alpha=0.3$ over the original distribution when training the SPM model. 
We use a shared vocabulary with 128K tokens for the 26 languages, and also append 32 special tokens (i.e., ``TBD0'' to ``TBD31'') for including extra tasks or data (e.g., tagged-BT~\cite{caswell2019tagged}).

\section{Approach}
\label{sec:approach}


\subsection{Data Augmentation}
\label{sec:data-augmentation}

We adopt data augmentation~(DA) to address the first challenge, i.e.,
``
{\bf The absence of training data for some to-be-evaluated language pairs}''.

Specifically, we use back-translation~(BT)~\cite{sennrich2016BT} and self-training~(ST)~\cite{jiao2020data,jiao2021self,jiao2022exploiting} to construct synthetic data.
However, previous study by \newcite{caswell2019tagged} suggests that the translationese issue in BT limits the performance, which can be mitigated with a special tag at source side~(i.e., tagged-BT).
To simplify the tagging procedure for the two opposite directions of each language pair, we use both BT and ST for each language pair~\cite{Wu2019ExploitMonoScale} and append a special token at the synthetic side of sentence pairs.
Formerly, for a language pair $(S,T)$ with the bitext data $\{\mathbf{x},\mathbf{y} \}$, the synthetic data by BT and ST will be $\{[\mathbf{x}';\mathtt{\langle DA\rangle}],\mathbf{y} \}$ and $\{\mathbf{x},[\mathbf{y}';\mathtt{\langle DA\rangle}] \}$, where $\mathtt{\langle DA\rangle}$ denotes the special tag for data augmentation.

We conduct data augmentation for both English-centric and non English-centric language pairs.
For English-centric language pairs, we randomly sample up to 1.0M English and non-English monolingual sentences from the training corpora for BT and ST, respectively. As for non English-centric language pairs, we translate the English side of English-centric pairs to non-English languages and construct up to 0.5M BT and ST sentence pairs, respectively.
Generally, the augmented data is included in \textbf{Large-234} to train the MNMT models.
However, the translation quality of those English-centric directions is also unreliable due to the limited data sizes, which may harm the performance of subsequent MNMT models. Besides, adding more synthetic data and language directions also slows down the convergence of the MNMT models.
Instead, we use the resulting MNMT models as backbones to finetune on the clean datasets.

\subsection{Distributionally Robust Optimization}
\label{sec:distributionally-robust-optimization}

We adopt the distributionally robust optimization~(DRO)~\cite{oren2019DROforLM,zhou2021DROforMNMT} technique to address the second challenge, i.e., ``{\bf The uneven optimization of language pairs due to data imbalance
}''.

Generally, temperature-based sampling~\citep{arivazhagan2019massively,conneau2020xlmr} is adopted to balance the training data across language pairs, which samples data from the smoothed data distribution as, $p_{\tau,i}=\frac{|D_i|^{1/\tau}}{\sum_j|D_j|^{1/\tau}}$.
This is equivalent to optimizing the re-weighted objective:
\begin{align}
    \mathcal{L}_{\tau}(\theta; D_{\rm train}) = \sum_{i\leq N} p_{\tau,i} \mathcal{L}(\theta; D_i),
\end{align}
where $|D_i|$ is the training data size of the $i$-th language pair, and $\tau$ denotes the temperature rate.
Obviously, $\tau = 1$ corresponds to the original data distribution while $\tau = \infty$ represents uniform sampling. In practice, $\tau > 1$ is adopted to oversample the low-resource language pairs, which significantly affects the results and needs to be tuned for different settings.  

Even if we can build a completely balanced dataset across language pairs, the varied task difficulty and cross-lingual similarity determine that the language pairs will still be optimized unevenly. DRO can address such a problem.
In contrast to temperature sampling which optimizes over a fixed training data distribution, DRO aims to find a model $\theta$ that can perform well on an entire set of potential test distributions, i.e., $\mathcal{U}(p^{\rm train})$, which is usually called \textit{uncertainty set}.
We adopt DRO with the $\chi^2$-uncertainty set introduced by \newcite{zhou2021DROforMNMT}, and reproduce the implementation for the practical many-to-many translation scenario.\footnote{The referred study only supports one-to-many and many-to-one translation scenarios on very small multilingual translation datasets.}
Similarly, we also incorporate the baseline losses calculated from a pretrained MNMT model to stabilize the training process of DRO.

\subsection{Language Family Grouping}
\label{sec:language-family-grouping}

We adopt language family grouping~(LFG) to alleviate the third challenge, i.e., ``
{\bf The curse of multilinguality}''~\cite{conneau2020xlmr}.

Specifically, we divide the target languages into 5 groups~( see Table~\ref{tab:language-family-grouping}) based on Table~\ref{tab:group-langpairs-1}. 
This is partially inspired by \newcite{eriguchi2022building}, which factorizes the many-to-many translation scenario (with $N \times N$ directions) into $N$ many-to-one scenarios by training a translation model for each. Since we have 26 languages involved in this shared task, factorizing the many-to-many scenario by the family of target languages is a more efficient choice. 
Since \textbf{swh} appears in both \textsc{HCEA} and \textsc{CA}, we include it in both Group-2 and Group-5 for training models. During inference, our scripts will automatically select the model of corresponding group according to the target language to be evaluated. Note that \textbf{swh} is only routed to Group-2 in inference.
\begin{table}[t!]
    \centering
    \caption{Language family grouping.}
    \vspace{-5pt}
    \begin{tabular}{c c }
    \toprule
        \bf Group & \multicolumn{1}{c}{\bf Target Languages}  \\
    \midrule
        1 & eng, fra \\
        2 & afr, nso, sna, ssw, tsn,  tso, xho, zul \\
        3 & amh, luo, orm, som, swh, wol \\
        4 & fuv, hau, ibo, yor \\
        5 & kam, kin, lin, lug, nya, swh, umb \\
    \bottomrule
    \end{tabular}
    \label{tab:language-family-grouping}
    \vspace{-5pt}
\end{table}

\begin{table*}[t!]
\centering
\caption{Evaluation results of our models on the devtest in terms of BLEU and ChrF++.}
\vspace{-5pt}
\resizebox{\textwidth}{!}{
\begin{tabular}{l ccccc c}
\toprule
\multirow{2}{*}{\bf Model} & \bf\textsc{X-Eng} & \bf\textsc{Eng-X} & \bf\textsc{X-Fra} & \bf\textsc{Fra-X} & \bf\textsc{X-X} & \bf\textsc{All}  \\
\cmidrule{2-7}
 & 22 & 22 & 4 & 4 & 48 & 100 \\
\midrule
\textsc{Transf-Deep} & 23.37/46.80 & 17.19/41.07 & 20.20/43.18 & 16.07/41.93 & 10.69/33.90  & 15.50/39.01 \\
\hdashline
\textsc{Borderline-Deep} & 25.87/48.83 & 18.24/42.05 & 22.31/44.91 & 16.74/42.32 & 11.66/34.89 & 16.86/40.23 \\
\textsc{Borderline-DWide} & 28.11/51.30  & 19.02/42.92 & 24.74/47.58 & 17.22/43.23 & 12.03/34.94 & 17.82/41.13 \\
\textsc{Borderline-DWide} w/ LFG &  28.26/51.37 & 19.38/43.37 &  24.87/47.74 & 17.48/43.72 & 12.04/35.01 & \bf 17.95/41.31 \\
\bottomrule
\end{tabular}
}
\label{tab:results-main-devtest}
\vspace{-2pt}
\end{table*}

\section{Experiments}

\subsection{Settings}
\label{sec:settings}

\paragraph{Model.}
We adopt the standard sequence-to-sequence Transformer~\cite{vaswani2017attention} as our architecture.
For the \textbf{Base-146} scale, we use a deep encoder of 24 layers and a relatively shallow decoder of 12 layers~\cite{Yang2021MicrosoftWMT}, with an embedding size of 1024, the feed-forward network size of 4096, and 16 attention heads (i.e., 0.59B parameters).
To stabilize the training of deep models, we follow~\newcite{wang2019DeepTransformer} to use pre-layer-normalization~(PLN) for both encoder and decoder layers.
For the \textbf{Large-234} scale, we enlarge the embedding size to 1536 to support more language pairs, which results in 1.02B parameters.
By default, we call these two models as \textsc{Transf-Deep} and \textsc{Transf-DWide}. The final models developed by our approaches are renamed as \textsc{Borderline-Deep} and \textsc{Borderline-DWide} for clarity.

\paragraph{Training.}
We train the MNMT models with the Adam optimizer~\cite{kingma2014adam} ($\beta_1=0.9, \beta_2=0.98$). The learning rate is set as 1e-4 with a warm-up step of 4000, followed by inverse square root decay. The models are trained with a dropout rate of 0.1 and a label smoothing rate of 0.1. All experiments are conducted on 32 NVIDIA A100 GPUs. 
Since the bitext data~($\approx$130M) for this year's shared task is less than 1/10 of that for last year's~($\approx$1.7B), we decide a batch size to be about 1/10 of that used in~\cite{Yang2021MicrosoftWMT}.
Specifically, we use 2048 max-tokens per GPUs and accumulate the gradients for every 8 steps to simulate the large batch size of 512K tokens. For language family grouping, we use the batch size of 131K tokens for each model.
For translation models trained by empirical risk minimization~(ERM) on the original training data, we upsample low-resource language pairs with the smoothing rate $\alpha = 0.3$~\cite{conneau2020xlmr}.
For those by DRO, we adopt the $\chi^2$-uncertainty set with the distribution divergence bounded by $\rho=0.1$. We use the ERM model to calculate the baseline losses for DRO.
We train these two kinds of models for at least 100K updates, upon which we may finetune for additional updates.

\paragraph{Evaluation.}
We use the dev and devtest of Flores-200 benchmark\footnote{\url{https://github.com/facebookresearch/flores/tree/main/flores200}} as our validation and test sets, and evaluate the MNMT models on the averaged last 10 checkpoints with sentencepiece BLEU and ChrF++. The sentencepiece model for evaluation also comes from the Flores-200 benchmark. 
The beam search process is performed with a beam size of 4 and a length penalty of 1.0.
Similar as the official competition results, we report our results by average-to-eng (\textsc{X-Eng}), average-from-eng (\textsc{Eng-X}), average-to-fra (\textsc{X-Fra}), average-from-fra (\textsc{Fra-X}), average-african-to-african (\textsc{X-X}), and the average for \textsc{All} translation directions.

\begin{table*}[t!]
\centering
\caption{Official evaluation results of submissions on the blind test sets in terms of BLEU and ChrF++.}
\vspace{-5pt}
\resizebox{0.84\textwidth}{!}{
\begin{tabular}{l ccccc c}
\toprule
\bf Submissions & \bf\textsc{X-Eng} & \bf\textsc{Eng-X} & \bf\textsc{X-Fra} & \bf\textsc{Fra-X} & \bf\textsc{X-X} & \bf\textsc{All}  \\
\cmidrule{2-7}
\textit{\#Lang-pairs} & 22 & 22 & 4 & 4 & 48 & 100 \\
\midrule
 & \multicolumn{6}{c}{\bf IIAI} \\
 \bf Primary & 23.15/43.88 & 12.80/37.52 & 18.35/41.08 & 13.08/38.70 & 2.58/19.52 & 10.40/30.47 \\
 \hdashline
 & \multicolumn{6}{c}{\bf GMU} \\
 \bf Language & 25.83/46.50 & 12.00/35.33 & 20.83/42.45 & 10.53/33.58 & 7.70/29.94 & 13.28/35.42 \\
 \bf Family & 25.88/46.55 & 11.98/35.30 & 20.73/42.30 & 10.75/34.03 & 7.68/29.92 & 13.28/35.42 \\
\midrule
 & \multicolumn{6}{c}{\bf Borderline (Ours)} \\
 \bf Contrastive & 25.84/47.46 & 13.85/39.05 & 21.00/44.10 & 13.85/39.58 & 8.03/30.93 & 13.98/37.23 \\
 \bf Primary & 26.05/47.56 & 14.06/39.53 & 21.13/44.05 & 14.05/40.10 & 8.04/31.04 & \bf 14.09/37.42 \\
\bottomrule
\end{tabular}
}
\label{tab:results-blind-test}
\vspace{-2pt}
\end{table*}

\subsection{Results}
\label{sec:results}

We list the evaluation results of our final models on the devtest in Table~\ref{tab:results-main-devtest}.
Both the baseline model \textsc{Transf-Deep} and our \textsc{Borderline-Deep} model are trained for 200K updates, while the two \textsc{Borderline-DWide} models are trained or finetuned for more than 300K updates. 

Generally, our models outperform the baseline \textsc{Transf-Deep} model significantly by up to +2.45 BLEU and +2.30 ChrF++ scores.
By looking into each category, we have some interesting findings: 
\begin{itemize}[leftmargin=10pt]
    \item By comparing \textsc{Borderline-Deep} and \textsc{Transf-Deep}, we find that the improvement on \textsc{X-Eng} is much larger than that on \textsc{Eng-X}. Similar phenomenon is also observed for \textsc{X-Fra} and \textsc{Fra-X}. It suggests that while DRO can achieve even improvement for one-to-many or many-to-one scenarios~\cite{zhou2021DROforMNMT}, it is heavily biased by the dominant languages (i.e., eng and fra) in the many-to-many scenario. 
    \item By comparing \textsc{Borderline-DWide} and \textsc{Borderline-Deep}, we find that enlarging the model capacity brings improvement to all categories but the most on \textsc{X-Eng} and \textsc{X-Fra}. It indicates that the \textit{curse of multilinguality} cannot be well solved by simply increasing model capacity as the most benefits are still occupied by the dominant languages (i.e., eng and fra).
    \item Language family grouping~(LFG) achieves more improvement on \textsc{Eng-X} and \textsc{Fra-X} than on the other categories, which confirms its effectiveness in alleviating the curse of multilingualty issue.
\end{itemize}

\begin{table}[t!]
\small
\setlength{\tabcolsep}{4pt}
\selectfont
\centering
\caption{Ablation study of our models with various strategies on the devtest. CT: continuous training; FT: finetuning; T-Enc: target language tags at encoder; LFG: language family grouping.}
\vspace{-5pt}
\begin{tabular}{c l c c r}
\toprule
\bf ID & \bf Model & \bf Step & \bf BLEU & \bf$\Delta$ \\
\midrule
\textcircled{\small{1}} & \textsc{Transf-Deep} & 100K & 15.03 & -/- \\
\textcircled{\small{2}}\cellcolor{blue!40} & ~~ + CT & 100K & 15.50 & +0.47 \\
\textcircled{\small{3}} & ~~ + FT on large-234 & 100K & 14.65 & -0.38 \\
\hdashline
\textcircled{\small{4}} & + DRO & 100K & 16.71 & +1.68 \\
\textcircled{\small{5}}\cellcolor{blue!40} & ~~ + CT & 100K & 16.86 & \bf +1.83 \\
\textcircled{\small{6}} & ~~ + T-Enc & 100K & 16.67 & +1.64  \\
\midrule
\textcircled{\small{7}} & \textsc{Transf-DWide} & 100K & 14.66 & -/- \\
\textcircled{\small{8}} & + DRO & 100K & 15.81 & +1.15 \\
\textcircled{\small{9}} & ~~ + FT on base-146 & 200K & 17.62 & +2.96 \\
\textcircled{\small{10}}\cellcolor{blue!40} & ~~~~ + FT on eval-106 & 50K & 17.82 & +3.16 \\
\textcircled{\small{11}}\cellcolor{blue!40} & ~~~~~~ + LFG & -/- & 17.95 & +3.29 \\
\bottomrule
\end{tabular}
\label{tab:results-devtest-ablation}
\vspace{-2pt}
\end{table}

\paragraph{Ablation Study.}
We present detailed ablation studies to investigate the effectiveness of various strategies, not only the three introduced in~\cref{sec:approach} but also some tricks. The results are listed in Table~\ref{tab:results-devtest-ablation}, where the lines marked in \colorbox{blue!40}{blue} (i.e., \textcircled{\small{2}}, \textcircled{\small{5}}, \textcircled{\small{10}} and \textcircled{\small{11}}) correspond to the four models in Table~\ref{tab:results-main-devtest}.
We list our observations as below:
\begin{itemize}[leftmargin=10pt]
    \item \textcircled{\small{3}} \textit{vs.}~\textcircled{\small{2}}: Directly finetuning the \textsc{Transf-Deep} model on the \textbf{Large-234} dataset induces the performance drop. One possible reason is that \textbf{Large-234} introduces much more translation directions, aggravating the \textit{curse of multilinguality} issue. Another reason is the low-quality data by data augmentation~(\cref{sec:data-augmentation}), which harms the optimization of models. Therefore, we only use \textbf{Large-234} to pretrain the \textsc{Transf-DWide} model and then finetune on the cleaner \textbf{Base-146} and \textbf{Eval-106} datasets.
    \item \textcircled{\small{6}} \textit{vs.}~\textcircled{\small{5}}: Previous studies~\cite{wang2022UncertZS} suggest that attaching target language tags at encoder (i.e., T-Enc) benefits the zero-shot translation performance, indicating a stronger cross-lingual transfer ability. However, we do not see any improvement of our models with T-Enc. The reason could be that, traditional studies on many-to-many translations are mainly conducted on the datasets with only one central language while we are now handling multiple central languages, making it a more complex scenario.
    \item \textcircled{\small{9}} \textit{vs.}~\textcircled{\small{10}} \textit{vs.}~\textcircled{\small{11}}: Finetuning on \textbf{Eval-106} slightly outperforms that on \textbf{Base-146} and the performance can be further improved with language family grouping. Obviously, as we reduce the language pairs involved in a single model, the \textit{curse of multilinguality} is alleviated.
\end{itemize}

\paragraph{Submissions.}
The \textsc{Borderline-DWide} and \textsc{Borderline-DWide}~w/~LFG models shown in Table~\ref{tab:results-main-devtest} (i.e., contrastive and primary versions) are submitted for official evaluation on the blind test sets. 
Table~\ref{tab:results-blind-test} summarizes the evaluation results of our submissions, where our models outperform the other teams' across all the evaluation groups. Finally, we achieve the \textbf{1st} place in this track.

\section{Conclusion}
\label{sec:conclusions}

In this paper, we describe Tencent's multilingual machine translation systems for the WMT22 shared task on Large-Scale Machine Translation Evaluation for African Languages.
We address three key challenges of this task by data augmentation, distributionally robust optimization~(DRO), and language family grouping, respectively, to develop our MNMT models.
Our submissions won the \textbf{1st} place in the \textbf{constrained} track. 
Extensive analyses also point out the drawbacks of larger models and DRO in addressing the curse of multilinguality, which warrants further research in the future.

\section*{Acknowledgments}
We sincerely thank the anonymous reviewers for their insightful suggestions on various aspects of this report.


\bibliographystyle{acl_natbib}
\bibliography{anthology,emnlp2022}

\end{document}